# LPQP for MAP: Putting LP Solvers to Better Use

**Patrick Pletscher**                                                    PLETSCHER@INF.ETHZ.CH
**Sharon Wulff**                                                 SHARON.WULFF@INF.ETHZ.CH
Department of Computer Science, ETH Zurich, Switzerland

## Abstract

MAP inference for general energy functions remains a challenging problem. While most efforts are channeled towards improving the linear programming (LP) based relaxation, this work is motivated by the quadratic programming (QP) relaxation. We propose a novel MAP relaxation that penalizes the Kullback-Leibler divergence between the LP pairwise auxiliary variables, and QP equivalent terms given by the product of the unaries. We develop two efficient algorithms based on variants of this relaxation. The algorithms minimize the non-convex objective using belief propagation and dual decomposition as building blocks. Experiments on synthetic and real-world data show that the solutions returned by our algorithms substantially improve over the LP relaxation.

## 1. Introduction

We study the problem of maximum a posteriori (MAP) inference in graphical models. The MAP task is to compute a minimal energy assignment of a set of dependent variables. This is a crucial problem in many applications such as computational biology, natural language processing and computer vision. In the general case, MAP inference is intractable, and therefore most of the current research efforts are concentrated on finding efficient and accurate approximation algorithms. In recent years, linear programming (LP) relaxations gained popularity due to their proven success in relevant applications. Several efficient algorithms have been developed to solve the linear program emerging from the relaxation. Despite their success, in many practical problems the solution attained by the LP relaxations is still far from the global minima.

Our work improves over the LP relaxation by leveraging on a second class of relaxations, namely the quadratic programming (QP) relaxation. The QP formulation offers a concise and compact description of the MAP problem. We formulate a joint LP and QP MAP objective, that encourages auxiliary variables present in the LP relaxation, to agree with their counterpart in the QP relaxation, through a penalty function. Despite of the non-convexity of this objective, we show that by slowly increasing the weight of the penalty, the solutions found are either competitive with, or in most cases better than the LP relaxation solutions. This is in general not the case for the few existing QP relaxation solvers.

We propose two variants of the penalty function, each leading to a different LPQP objective. We show that the resulting non-convex objectives can be decomposed into a difference of convex functions, which we solve using the convex-concave procedure (CCCP). Having tackled the non-convexity with the CCCP, we solve one of the remaining convex problems with the dual decomposition method, and show that the other can be addressed with the norm-product belief propagation. Interestingly, the main computational task of both of the resulting LPQP algorithms, turns out to be solving known entropy-augmented LPs.

Our contributions are as follows: First we introduce a combined LPQP objective, incorporating the QP constraints through a soft penalty function in the objective. We propose two alternatives for the penalty function, which differ in the way the edges in the graph are weighted. Secondly, we derive CCCP based algorithms for the LPQP objectives, and show that their core computational effort reduces to current entropy-augmented LP solvers. This demonstrates that these modern LP solvers can in some cases be utilized in a better way, leading to possibly faster convergence, as well as lower energy MAP solutions. Through experiments on various datasets, we demonstrate the performance of the suggested LPQP MAP inference in comparison to other commonly used solvers.

## 2. Background and Notation

For an undirected graph $\mathcal{G} = (\mathcal{V}, \mathcal{E})$, the MAP problem is to assign each node in the graph to a class or category, such that the overall assignment minimizes an associated energy. Let $x_i$ denote a discrete variable with a finite domain $\mathcal{X}_i$[1], representing the assignment of the $i$-th node. The MAP problem is defined as

$$\min_{\boldsymbol{x}} \sum_{i \in \mathcal{V}} \theta_i(x_i) + \sum_{(i,j) \in \mathcal{E}} \theta_{ij}(x_i, x_j). \quad (1)$$

Where $\theta_i(x_i)$ and $\theta_{ij}(x_i, x_j)$ are unary and pairwise potential functions associated with the node and edge assignments. Problem (1) can be expressed as an integer quadratic program using a $K$-ary coding:

$$\min_{\boldsymbol{\mu}} \quad \sum_{i \in \mathcal{V}} \boldsymbol{\theta}_i^{\mathsf{T}} \boldsymbol{\mu}_i + \sum_{(i,j) \in \mathcal{E}} \boldsymbol{\mu}_i^{\mathsf{T}} \boldsymbol{\Theta}_{ij} \boldsymbol{\mu}_j \quad (2)$$

$$\text{s.t.} \quad \mu_{i;k} \in \{0, 1\} \quad \forall i, k \quad \text{and} \quad \sum_k \mu_{i;k} = 1 \quad \forall i.$$

The pairwise and unary potentials in (2), are represented as a matrix $\boldsymbol{\Theta}_{ij}$ and a vector $\boldsymbol{\theta}_i$, respectively.

Variational approaches to MAP inference reformulate the combinatorial optimization problem in (1) as a continuous optimization problem. The next sections formally define two such approaches, namely the LP and QP relaxations. In general, the LP minimization results in a lower bound on the energy of the global minimizer, while the QP results in an upper bound.

### 2.1. Linear Programming Relaxation

The LP approach (Schlesinger, 1976; Wainwright & Jordan, 2008) is based on a convex relaxation of (2), where an additional variable $\boldsymbol{\mu}_{ij}$ is included for each edge. Proper local marginalization is enforced through summation constraints. The LP reads as

$$\min_{\boldsymbol{\mu} \in \mathcal{L}_{\mathcal{G}}} \sum_{i \in \mathcal{V}} \boldsymbol{\theta}_i^{\mathsf{T}} \boldsymbol{\mu}_i + \sum_{(i,j) \in \mathcal{E}} \boldsymbol{\theta}_{ij}^{\mathsf{T}} \boldsymbol{\mu}_{ij}, \quad (3)$$

with $\mathcal{L}_{\mathcal{G}}$, the local marginal polytope:

$$\mathcal{L}_{\mathcal{G}} = \left\{ \boldsymbol{\mu} \;\middle|\; \begin{array}{l} \sum_k \mu_{i;k} = 1 \quad \forall i \in \mathcal{V} \\ \sum_l \mu_{ij;kl} = \mu_{i;k} \quad \forall k, (i,j) \in \mathcal{E} \\ \sum_k \mu_{ij;kl} = \mu_{j;l} \quad \forall l, (i,j) \in \mathcal{E} \\ \mu_{ij;kl} \geq 0 \quad \forall k, l, (i,j) \in \mathcal{E} \end{array} \right\}.$$

In the general case, $\mathcal{L}_{\mathcal{G}}$ is an inexact description of the so called marginal polytope $\mathcal{M}_{\mathcal{G}}$, which requires an exponentially large number of constraints (Wainwright

& Jordan, 2008). If $\mathcal{L}_{\mathcal{G}}$ in (3) is replaced by $\mathcal{M}_{\mathcal{G}}$, then the solution recovers the true MAP assignment. A solution to an LP-based approach, admits an easy to verify certificate of optimality. If the solution is integer, it is the global optimum.

The work in (Sontag et al., 2008) proposes to tighten the polytope by including summation constraints over larger subsets of variables. This approach has been successful in identifying the global minima for some problems. However, it suffers from an increased complexity as ultimately an exponentially large set of possible constraints might need to be searched over.

### 2.2. Quadratic Programming Relaxation

An alternative relaxation of the integer quadratic program in (2) is obtained by simply dropping the integer constraints. The resulting QP is given by:

$$\min_{\boldsymbol{\mu}} \quad \sum_{i \in \mathcal{V}} \boldsymbol{\theta}_i^{\mathsf{T}} \boldsymbol{\mu}_i + \sum_{(i,j) \in \mathcal{E}} \boldsymbol{\mu}_i^{\mathsf{T}} \boldsymbol{\Theta}_{ij} \boldsymbol{\mu}_j \quad (4)$$

$$\text{s.t.} \quad 0 \leq \mu_{i;k} \leq 1 \quad \forall i, k \quad \text{and} \quad \sum_k \mu_{i;k} = 1 \quad \forall i.$$

A major advantage of the QP relaxation, is the fact that it is tight. In this context the tightness means that the minimizer of (4) is also the minimizer of (1), as was shown in (Ravikumar & Lafferty, 2006). The QP also benefits from a more compact description compared to the LP relaxation, as it requires fewer constraints and variables to formulate the *exact* MAP problem. The variable vector $\boldsymbol{\mu}$, is of size $K \cdot |\mathcal{V}| + K^2 \cdot |\mathcal{E}|$ in the LP (3) and only $K \cdot |\mathcal{V}|$ in the QP. The biggest drawback of the QP relaxation, is that in the general case the optimization problem is non-convex due to the edges product term. This fact renders an exact minimization difficult.

In terms of motivation, our work is similar to the QP relaxation approach. The QP formulation of the MAP problem (4) was introduced in (Ravikumar & Lafferty, 2006), but stems from classical mean-field approaches. Ravikumar & Lafferty (2006) solved the non-convex problem using a convex relaxation. The solution was later improved in (Kappes & Schnoerr, 2008) through a difference of convex functions formulation. Both solvers are generic in the sense that they do not exploit the graph structure. Recently Kumar & Zilberstein (2011) introduced a message-passing algorithm for solving the QP relaxation. While improving the run time over the other two algorithms, it still generally suffers from poor solutions due to local minima. The QP solvers often deal with this drawback by restarting with different initializations. We observed that our LPQP algorithms, are much more resilient with respect to the initialization. In all of the experiments we conducted, a restart was never required. We attribute this behavior to the gradual progression between the LP and QP. Finally, in concurrent work Kumar et al. (2012) propose a hybrid LP and QP approach to MAP, similar to our formulation discussed in the next section. The resulting optimization problem is solved by a custom message-passing scheme. Our work on the other hand, in its essence reduces to well-known entropy-augmented LP objectives, for which efficient message-passing algorithms exist.

---

[1] For notational convenience we assume $\mathcal{X}_i = \{1, \ldots, K\}$, in the experiments we will however also consider settings where the domain of the variables has different size.

## 3. Combined LP and QP Relaxation

We propose to optimize an objective which is a combination of the LP and QP relaxations. We retain the auxiliary variables $\boldsymbol{\mu}_{ij}$ of the pairwise terms, but force these variables to agree with the product of the unary marginals $\boldsymbol{\mu}_i$ and $\boldsymbol{\mu}_j$. The constraints, given by $\mathrm{vec}(\boldsymbol{\mu}_i\boldsymbol{\mu}_j^\mathsf{T}) = \boldsymbol{\mu}_{ij} \ \forall (i,j) \in \mathcal{E}^2$, are enforced through a penalty function $g(\cdot)$ incorporated in the objective. The extent to which the constraint is enforced, is regulated by the parameter $\rho$.

We focus on the Kullback-Leibler (KL) divergence as the penalty function, due to the probabilistic nature of the compared marginal terms. For probability distributions $\boldsymbol{p}$ and $\boldsymbol{q}$ of a discrete random variable, their KL divergence is defined to be

$$D_{KL}(\boldsymbol{p}, \boldsymbol{q}) := \sum_k p_k \log\left(\frac{p_k}{q_k}\right).$$

The general form of the combined objective reads as

$$\min_{\boldsymbol{\mu} \in \mathcal{L}_{\mathcal{G}}} \boldsymbol{\theta}^\mathsf{T}\boldsymbol{\mu} + \rho g(\boldsymbol{\mu}). \qquad (5)$$

The first term is simply the LP objective (3), written as a scalar product between the potential function, and the concatenated unary and pairwise variables. We investigate two constructions of the penalty term. The constructions differ in the weighting of the edges.

**Uniform Weighting** The KL divergence is penalized in the same way for all the edges in the graph:

$$g^{uni}(\boldsymbol{\mu}) := \sum_{(i,j) \in \mathcal{E}} D_{KL}(\boldsymbol{\mu}_{ij}, \mathrm{vec}(\boldsymbol{\mu}_i\boldsymbol{\mu}_j^\mathsf{T})). \qquad (6)$$

**Tree-based Weighting** The KL divergence is penalized uniformly within a forest-shaped sub-graph:

$$g^{tree}(\boldsymbol{\mu}) := \sum_{a \in \mathcal{A}} \eta_a \left( \sum_{(i,j) \in \mathcal{E}_a} D_{KL}(\boldsymbol{\mu}_{ij}, \mathrm{vec}(\boldsymbol{\mu}_i\boldsymbol{\mu}_j^\mathsf{T})) \right). \qquad (7)$$

---

[2]Here $\mathrm{vec}(\boldsymbol{\mu}_i\boldsymbol{\mu}_j^\mathsf{T})$ denotes the vectorized version of the outer product of $\boldsymbol{\mu}_i$ and $\boldsymbol{\mu}_j$.

We assume that a decomposition of the original graph into acyclic subgraphs exists, and is given by

$$\mathcal{G}_a = (\mathcal{V}_a, \mathcal{E}_a), \qquad \mathcal{V} = \bigcup_{a \in \mathcal{A}} \mathcal{V}_a, \qquad \mathcal{E} = \bigcup_{a \in \mathcal{A}} \mathcal{E}_a.$$

The positive weights $\eta_a$ are tree specific, and assumed to sum to one. In this work we simply used $\eta_a = 1/|\mathcal{A}|$.

For $\rho = 0$, (5) amounts to the standard LP relaxation. On the other extreme when $\rho \to \infty$, the constraints $\mathrm{vec}(\boldsymbol{\mu}_i\boldsymbol{\mu}_j^\mathsf{T}) = \boldsymbol{\mu}_{ij} \ \forall (i,j) \in \mathcal{E}$ are fulfilled and the QP relaxation is recovered. By successively increasing $\rho$ during the run of our algorithms, we achieve a gradual enforcement of the constraints.

## 4. LPQP Algorithms

In this section we derive two algorithms for the non-convex LPQP objective in (5), with the different penalty terms in (6) and (7).

### 4.1. Difference of Convex Functions (DC)

The *convex-concave procedure* (CCCP) (Yuille & Rangarajan, 2003), can be applied to a constrained optimization problem, where the objective is non-convex, provided that the objective has a decomposition into a convex and a concave part. In our setting, we wish to find a decomposition of the form

$$\min_{\boldsymbol{\mu} \in \mathcal{L}_{\mathcal{G}}} u_\rho(\boldsymbol{\mu}) - v_\rho(\boldsymbol{\mu}),$$

where both, $u_\rho(\boldsymbol{\mu})$ and $v_\rho(\boldsymbol{\mu})$ are convex. The CCCP algorithm proceeds by iteratively solving a convexified objective, obtained by a linearization of $v_\rho(\boldsymbol{\mu})$:

$$\boldsymbol{\mu}^{t+1} = \operatorname*{argmin}_{\boldsymbol{\mu} \in \mathcal{L}_{\mathcal{G}}} u_\rho(\boldsymbol{\mu}) - \boldsymbol{\mu}^\mathsf{T}\nabla v_\rho(\boldsymbol{\mu}^t). \qquad (8)$$

The decompositions of the two objectives, as well as the gradients of the concave part, are shown in Figure 1. In the derivations we used the following identity

$$D_{KL}(\boldsymbol{\mu}_{ij}, \mathrm{vec}(\boldsymbol{\mu}_i\boldsymbol{\mu}_j^\mathsf{T})) = H(\boldsymbol{\mu}_i) + H(\boldsymbol{\mu}_j) - H(\boldsymbol{\mu}_{ij}),$$

which holds due to the marginalization constraints of the pairwise marginals (Wainwright & Jordan, 2008).

For both objectives, the convex part $u_\rho(\boldsymbol{\mu})$ consists of the original LP formulation, with an additional term that encourages configurations with a large entropy. This term in the uniform weights penalty, is the entropy of the pairwise marginals, whereas in the tree-based penalty, it is the sum of tree entropies.

The concave part of the decompositions, $v_\rho$, corresponds to an entropy of the unary marginals. In the

Uniform weighting        Tree-based weighting

$$u_\rho(\boldsymbol{\mu}) = \boldsymbol{\theta}^\mathsf{T}\boldsymbol{\mu} - \rho \sum_{(i,j)\in\mathcal{E}} H(\boldsymbol{\mu}_{ij})$$    $$u_\rho(\boldsymbol{\mu}) = \boldsymbol{\theta}^\mathsf{T}\boldsymbol{\mu} - \rho \sum_{a\in\mathcal{A}} \eta_a \left( \sum_{(i,j)\in\mathcal{E}_a} H(\boldsymbol{\mu}_{ij}) - \sum_{i\in\mathcal{V}_a} (d_i^a - 1) H(\boldsymbol{\mu}_i) \right)$$

$$v_\rho(\boldsymbol{\mu}) = -\rho \sum_{i\in\mathcal{V}} d_i H(\boldsymbol{\mu}_i)$$    $$v_\rho(\boldsymbol{\mu}) = -\rho \sum_{a\in\mathcal{A}} \eta_a \sum_{i\in\mathcal{V}_a} H(\boldsymbol{\mu}_i)$$

$$\frac{\partial v_\rho(\boldsymbol{\mu})}{\partial \mu_{i;k}} = \rho d_i (1 + \log \mu_{i;k})$$    $$\frac{\partial v_\rho(\boldsymbol{\mu})}{\partial \mu_{i;k}} = \rho \sum_{a\in\mathcal{A}(i)} \eta_a (1 + \log \mu_{i;k})$$

*Figure 1.* Difference of convex function decomposition of the combined LPQP objective in (5) for the two different penalty terms. *Left:* decomposition for the uniform penalty term in (6). Right: decomposition for the tree weighted penalty term in (7). In both cases the derivative w.r.t. the pairwise marginal variables is zero. Here $d_i$ denotes the degree of the $i$-th node in the graph, and $d_i^a$ its degree in the sub-graph indexed by $a$. $\mathcal{A}(i)$ denotes the set of all trees that contain node $i$. For a distribution $\boldsymbol{p}$ of a discrete random variable, $H(\boldsymbol{p})$ denotes the entropy, $H(\boldsymbol{p}) := -\sum_k p_k \log(p_k)$.

CCCP step (8), $\log(\boldsymbol{\mu}_i)$ is replaced by $\log(\boldsymbol{\mu}_i^t)$, the marginal from the previous iteration, resulting in an entropy approximation.

### 4.2. Algorithm Overview

The general scheme of the suggested LPQP algorithms is shown in Algorithm 1. The algorithm consists of two loops. The inner loop solves the DC problem for a fixed penalty parameter $\rho$, whereas the outer loop gradually increases the value of $\rho$.

---

**Algorithm 1** LPQP algorithm scheme for MAP.

**Require:** $\mathcal{G} = (\mathcal{V}, \mathcal{E})$, $\boldsymbol{\theta}$.
1: initialize $\boldsymbol{\mu} \in \mathcal{L}_\mathcal{G}$ uniform, $\rho = \rho_0$.
2: **repeat**
3:   $t = 0, \boldsymbol{\mu}^0 = \boldsymbol{\mu}$.
4:   **repeat**
5:     $\boldsymbol{\mu}^{t+1} = \operatorname{argmin}_{\boldsymbol{\tau}\in\mathcal{L}_\mathcal{G}} u_\rho(\boldsymbol{\tau}) - \boldsymbol{\tau}^\mathsf{T}\nabla v_\rho(\boldsymbol{\mu}^t)$.
6:     $t = t + 1$.
7:   **until** $\|\boldsymbol{\mu}^t - \boldsymbol{\mu}^{t-1}\|_2 \le \epsilon_{\mathrm{dc}}$.
8:   $\boldsymbol{\mu} = \boldsymbol{\mu}^t$.
9:   increase $\rho$.
10: **until** $\|\boldsymbol{\mu} - \boldsymbol{\mu}^0\|_2 \le \epsilon_\rho$.
11: **return** $\boldsymbol{\mu}$.

---

The main computational task is in line 5, where a particular instance of a convex optimization problem is solved. Warm-starting the problem in line 5 with the previous solution between successive calls, leads to a substantial speed-up. We choose the initial $\rho = \rho_0$ depending on the scaling of the energies, and use a multiplicative increase with a fixed value. In the experiments we use a multiplicative factor of 1.5, but the results were not very sensitive to this choice.

**Solution Rounding** Similarly to the LP and QP relaxations, the solutions returned by the LPQP algorithms can be fractional. Since the LPQP scheme ultimately solves a variant of the QP relaxation, to attain the final integer solutions, we use the QP solution rounding scheme suggested in (Ravikumar & Lafferty, 2006). Given a unary marginals vector $\boldsymbol{\mu}^*$, we assign the $i$-th node the label $x_i^*$ given by

$$x_i^* = \operatorname*{argmin}_k \left( \theta_{i;k} + \sum_{j\in\mathcal{N}(i)} \sum_l \theta_{i,j;k,l} \mu_{j;l}^* \right).$$

Here $\mathcal{N}(i)$ denotes the neighbors of node $i$. After determining the label of the $i$-th variable, we set $\mu_{i;x_i^*}^* = 1$ and $\mu_{i;k}^* = 0 \ \forall \ k \neq x_i^*$, and continue until labels are assigned to all nodes. It can be verified that the rounded solution has an energy that is smaller or equal to the energy of the initial solution $\boldsymbol{\mu}^*$.

### 4.3. Uniform Weighting

The convex sub-problem we get in the CCCP step with the uniform weighting penalty function (6), is given by

$$\min_{\boldsymbol{\mu}\in\mathcal{L}_\mathcal{G}} \sum_{i\in\mathcal{V}} \tilde{\boldsymbol{\theta}}_i^\mathsf{T}\boldsymbol{\mu}_i + \sum_{(i,j)\in\mathcal{E}} \boldsymbol{\theta}_{ij}^\mathsf{T}\boldsymbol{\mu}_{ij} - \rho \sum_{(i,j)\in\mathcal{E}} H(\boldsymbol{\mu}_{ij}). \quad (9)$$

where $\tilde{\boldsymbol{\theta}}_i$, is a modification of the unary potentials by an additional gradient term, originating in the linearized part of the DC decomposition (8) [3].

$$\tilde{\boldsymbol{\theta}}_i = \boldsymbol{\theta}_i - \rho d_i \log(\boldsymbol{\mu}_i^t), \quad (10)$$

---
[3] The $\rho d_i$ term in $\nabla v_\rho$ is constant and can therefore be dropped.

As a result of this unary potentials modification, configurations with small probability in the previous iteration $t$, are vigorously dis-encouraged.

**Belief Propagation** The convex problem in (9) is solved by the norm-product belief-propagation (BP) algorithm (Hazan & Shashua, 2010). It is a primal-dual ascent algorithm and is guaranteed to converge to the global optimum for any choice of $\rho > 0$.

The norm-product algorithm applied to (9) computes messages passed from node $j$ to node $i$ as follows

$$m_{j \to i}(x_i) \propto$$
$$\left( \sum_{x_j} \psi_{ij}^{1/\rho}(x_i, x_j) \frac{\psi_j^{1/(d_j\rho)}(x_j) \prod_{s \in \mathcal{N}(j)} m_{s \to j}^{1/(d_j\rho)}(x_j)}{m_{i \to j}^{1/\rho}(x_j)} \right)^{\rho}$$

here we define $\psi_{ij}(x_i, x_j) = \exp(-\theta_{ij}(x_i, x_j))$ and $\psi_i(x_i) = \exp(-\tilde{\theta}_i(x_i))$. Upon convergence the marginals $\boldsymbol{\mu}_i$ are obtained by multiplying the incoming messages at variable $i$:

$$\mu_i(x_i) \propto \left( \psi_i(x_i) \prod_{j \in \mathcal{N}(i)} m_{j \to i}(x_i) \right)^{1/(d_i\rho)}.$$

Due to warm starting with the previous DC iteration solution, typically only few passes through the graph are needed for the messages to converge in the later stages of the run.

### 4.4. Tree-based Weighting

The convex sub-problem corresponding to the CCCP step with the tree-based weighting penalty (7) is,

$$\min_{\boldsymbol{\mu} \in \mathcal{L}_\mathcal{G}} \sum_{i \in \mathcal{V}} \tilde{\boldsymbol{\theta}}_i^\mathsf{T} \boldsymbol{\mu}_i + \sum_{(i,j) \in \mathcal{E}} \boldsymbol{\theta}_{ij}^\mathsf{T} \boldsymbol{\mu}_{ij} - \rho \sum_{a \in \mathcal{A}} \eta_a H_{\text{tree}}^a(\boldsymbol{\mu}). \quad (11)$$

Where we define the entropy of a tree by

$$H_{\text{tree}}^a(\boldsymbol{\mu}) := \left( \sum_{(i,j) \in \mathcal{E}_a} H(\boldsymbol{\mu}_{ij}) - \sum_{i \in \mathcal{V}_a} (d_i^a - 1) H(\boldsymbol{\mu}_i) \right).$$

As before, the linearization of the concave part in the CCCP step, results in a modification of the unaries

$$\tilde{\boldsymbol{\theta}}_i = \boldsymbol{\theta}_i - \rho \sum_{a \in \mathcal{A}(i)} \eta_a \log(\boldsymbol{\mu}_i^t). \quad (12)$$

**Dual Decomposition** The dual decomposition framework (Bertsekas, 1999; Komodakis et al., 2007), can be applied to an optimization problem provided that the objective can be decomposed into several sub-problems, also known in the literature as the slave problems. The global variables, $\boldsymbol{\mu}$ in our case, are replaced with local copies in each slave problem, denoted here by $\boldsymbol{\nu}^a$, such that the minimization of the slave problems can be carried out independently. To enforce the local variables corresponding to the same original variables to assume the same value, a designated constraint is introduced. The optimization of the sum of slave problems, subject to these constraints, is called the master problem. A dual decomposition of problem (11), was carried out in (Domke, 2011). We use the same decomposition, but take a different route optimizing the resulting master problem.

$$\min_{\boldsymbol{\mu} \in \mathcal{L}_\mathcal{G}} \sum_{a \in \mathcal{A}} \min_{\boldsymbol{\nu}^a \in \mathcal{L}_{\mathcal{G}_a}} s_a(\boldsymbol{\nu}^a) \quad (13)$$
$$\text{s.t.} \quad \boldsymbol{\nu}_i^a = \boldsymbol{\mu}_i \quad \forall i, a \in \mathcal{A}.$$
$$\boldsymbol{\nu}_{ij}^a = \boldsymbol{\mu}_{ij} \quad \forall (i,j), a \in \mathcal{A}.$$

Where the slave problems are defined as

$$s_a(\boldsymbol{\nu}) := \sum_{i \in \mathcal{V}_a} \bar{\boldsymbol{\theta}}_i^\mathsf{T} \boldsymbol{\nu}_i + \sum_{(i,j) \in \mathcal{E}_a} \bar{\boldsymbol{\theta}}_{ij}^\mathsf{T} \boldsymbol{\nu}_{ij} - \rho \eta_a H_{\text{tree}}^a(\boldsymbol{\nu}).$$

Note that since the summation over the trees now extends to include the unary and pairwise terms, the corresponding potentials should be adjusted accordingly

$$\bar{\boldsymbol{\theta}}_i = \frac{\tilde{\boldsymbol{\theta}}_i}{|\mathcal{A}(i)|}, \qquad \bar{\boldsymbol{\theta}}_{ij} = \frac{\boldsymbol{\theta}_{ij}}{|\mathcal{A}(i,j)|}.$$

Each slave problem is defined over a tree structured graph and can therefore be solved exactly using the sum-product algorithm, in two passes over the tree. The temperature in this case is $\rho \eta_a$. We solve the dual of the master problem using the FISTA (Beck & Teboulle, 2009) algorithm. A similar solution was demonstrated in (Savchynskyy et al., 2011), on a more restricted decomposition. We refer to the appendix for more technical details.

### 4.5. Entropy-augmented LP Solvers

Recently, several works (Jojic et al., 2010; Savchynskyy et al., 2011) proposed to smooth the LP objective by adding a term that favors entropic marginals. The merit of this additional term is in overcoming the non-smoothness of the objective. In order to ultimately solve the original LP, these entropy-augmented solvers progressively lower the entropy term. Naturally, the convergence of these algorithms is fairly fast in the beginning. This line of research originates in Nesterov's work on fast gradient methods (Nesterov, 1983).

The proposed LPQP solvers have the opposite behavior with respect to the smoothness of the objective.

The influence of the entropy term is rather increased through the progression of the algorithm, leading to favorable convergence properties.

## 5. Experiments

We use LPQP-U to refer to the implementation of the uniform weighting penalty, and LPQP-T for the tree-based weighting. In the experiments where the graph did not have a natural decomposition, we used a depth-first search algorithm to construct a tree decomposition in a greedy fashion for LPQP-T.

**Benchmarked Methods** We compare the performance of LPQP-U and LPQP-T with the widely used MAP algorithms, tree-reweighted belief propagation (TRWS) (Kolmogorov, 2006) and max-product LP (MPLP) (Sontag et al., 2008), both of which are LP relaxations. For both algorithms we used the implementation made available by the authors. These algorithms represent different trade-offs in performance. TRWS is a highly efficient message-passing algorithm for the standard LP relaxation. It is much faster than the MPLP, especially on large instances where the MPLP convergence is pretty slow. MPLP on the other hand, initially solves the LP relaxation over the local polytope, and in later iterations includes additional summation constraints over sets of three or four variables. This strategy naturally leads to lower (better) energy solutions, on instances where the LP relaxation is not tight. The MPLP was shown to identify the global optimum for some problems.

**Performance Measures** In this work we mainly compared the quality of the solutions, which in the MAP setting is most naturally measured by the energy associated with an assignment (1). Strictly comparing energy values is problematic for two reasons. The values lack proper scaling required for quantitative comparison of different results on the same problem instance, and are not comparable across instances. We therefore exercise the following scoring procedure. Let $e_1, \ldots, e_J$ denote the energies of the compared solutions, we set

$$s_i = \frac{\max_{1 \leq j \leq J}(e_j) - e_i}{\max_{1 \leq j \leq J}(e_j) - \min_{1 \leq j \leq J}(e_j)} \qquad (14)$$

as the score of the $i$-th method. This scheme assigns the worst and the best methods, scores of zero and one respectively. The remaining methods get a fraction relative to their value between the best and the worst result. This procedure is not flawless since the scores are still computed relative to the worst energies. It was most often the case though, that TRWS was the lowest scoring method. Being an often used algorithm with provable merits, using it as a normalizing measure is in our opinion a sensible choice. In experiments where the optimal value is known, we use this value instead of $\min_{1 \leq j \leq J} e_j$. In addition to comparing the quality of the solution, we comment about the trends in the efficiency (run-time) of the various methods.

### 5.1. Synthetic Potts Model Data

We follow a similar experimental setup as in (Ravikumar et al., 2010). The graph is a 4-nearest neighbor grid of varying size. We used $M = 60, 90, 120$ where $M$ is the grid side-length, and $M^2$ is the overall number of variables. We used $K = 2$ and $K = 5$ for the number of states. The unary potentials were randomly set to $\theta_{i;k}(x_i) \sim Uniform(-\sigma, \sigma)$, and for $\sigma$ we used values in $[0.05, 0.5]$. Note that the problem instance gets harder for small values of $\sigma$, this parameter can be understood as the signal-to-noise ratio. The pairwise potentials $\theta_{ij}(x_i, x_j)$, were set to penalize agreements or disagreements of the labels, by an amount $\alpha_{ij} \sim Uniform(-1, 1)$, chosen at random. We set $\theta_{ij}(x_i, x_j) = 0$ if $x_i \neq x_j$ and $\alpha_{ij}$ otherwise. In this experiment we choose the graph decomposition for the LPQP-T solution as the vertical and horizontal split of the grid edges. The two trees have all the original nodes in common, but no overlapping edges.

The results of the comparison using the performance measure given in (14), are presented in Table 1. For each choice of parameters, we averaged the scores of 5 runs. Furthermore, Figure 2 shows the progress of the objective during a run of the LPQP-U algorithm.

| $M$ (size) | 60 | | 90 | | 120 | |
|---|---|---|---|---|---|---|
| $K$ (# states) | 2 | 5 | 2 | 5 | 2 | 5 |
| | | | $\sigma = 0.05$ | | | |
| MPLP | 0.71 | 0.99 | 0.51 | 0.96 | 0 | 0.95 |
| LPQP-U | 0.97 | 0.99 | 0.97 | 1 | 0.98 | 1 |
| LPQP-T | 1 | 0.97 | 1 | 0.98 | 1 | 0.98 |
| TRWS | 0 | 0 | 0 | 0 | 0.39 | 0 |
| | | | $\sigma = 0.5$ | | | |
| MPLP | 1 | 1 | 1 | 1 | 1 | 0.99 |
| LPQP-U | 0.99 | 0.92 | 0.99 | 0.91 | 1 | 0.94 |
| LPQP-T | 0.99 | 0.95 | 0.99 | 0.94 | 0.99 | 0.96 |
| TRWS | 0 | 0 | 0 | 0 | 0 | 0 |

*Table 1.* Averaged scores achieved by the MAP solvers on the synthetic grid data. The scores, computed according to (14), assign in each run 1 and 0 to the best and the worst objective values. The remaining algorithms get a fractional score reflecting their relative objective value.

In terms of running time, TRWS was always first to output a solution, followed by the LPQP algorithms. MPLP was always slower and on the larger instances did not converge within a predefined maximal time. We therefore restricted the number of tightening iterations of MPLP to a maximum of 1000. A tightening iteration includes additional constraints into the local marginal polytope. Even after this change, MPLP was still considerably slower than the other algorithms. Between the LPQP algorithms, the LPQP-U was most often faster than LPQP-T.

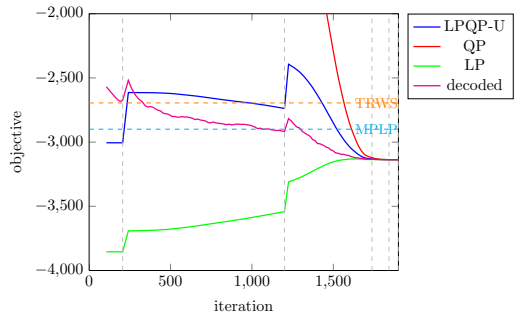

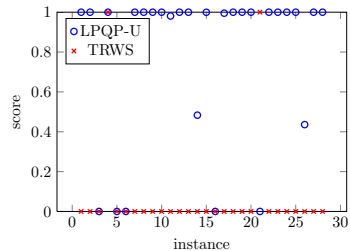

Figure 2. Development of the different objectives (for the same $\mu$) during a run of the LPQP-U. The decoded objective refers to the current solution independently rounded to integer values. The vertical lines show iterations where $\rho$ was increased. The horizontal lines show the energy of the solution found by TRWS and MPLP, respectively.

Figure 3. Protein prediction results for instances where the LP is not tight. LPQP-U improves on TRWS in all but one cases. For 20 of the 28 instances LPQP finds the true MAP.

As we expect, TRWS returned the worst assignment on almost all configurations. The energies obtained by LPQP-U, LPQP-T and MPLP were in general very close. We observe that both of the LPQP algorithms, returned slightly better solutions in comparison to the MPLP, when the potentials were sampled with lower signal-to-noise ratio $\sigma$.

The run time of LPQP-T seems to be mostly influenced by the structure of the decomposition. In later experiments where the decomposition consisted of a larger number of trees with more variables in common, the LPQP-T was significantly slower compared to the LPQP-U. In terms of the energy of the solutions, the two algorithms were very similar. For this reason we report from now on the LPQP-U only. The LPQP-T can still be beneficial in settings where the computations are done on a distributed system.

### 5.2. Protein Design & Prediction

The protein inference problem discussed in (Yanover et al., 2006), consists of two tasks: protein side-chain prediction and protein design. For the protein prediction task, it was shown in (Yanover et al., 2006) that only 30 out of the 370 protein prediction instances, the LP relaxation is not tight. For 28 of them, the true MAP was computed using general integer pro-gramming techniques, Figure 3 visualizes the results on these instances. The LPQP found the global minimum of roughly 2/3 of these more difficult instances. On the remaining 340 instances, the LP is tight. The LPQP found the global optimum in all but three cases (results are not shown). MPLP was applied to this task in (Sontag et al., 2008), and achieved the global optimum on all instances.

The protein design task consists of 97 instances. We used MPLP to compute the global optimum, but for one of the instances, MPLP did not finish within a time-budget of 7 days. The average scores for the remaining 96 instances are as follows. LPQP-U: 0.93, MPLP: 1 and TRWS: 0.03. The average energies are: LPQP-U: $-184.06$, MPLP: $-184.60$, TRWS: $-173.55$. The QP message-passing algorithm in (Kumar & Zilberstein, 2011), was tested on this task as well. The evaluation criteria used in this work was the average (across the 97 instances) percentage of the optimal value. While the reported average value in (Kumar & Zilberstein, 2011) is 97.7%, our solution achieves 99.7% percentage of the optimal value on average.

### 5.3. Decision Tree Fields

As a last experiment we apply our LPQP algorithm to the recently published "hard discrete energy minimization instances" dataset (Nowozin et al., 2011), available on the authors webpage. The task is to fill in, or inpaint, a blanked out area in a binary image of Chinese handwritten characters, see Figure 4. The dataset consists of 100 energy minimization instances, and comes with approximate MAP solutions obtained using simulated annealing (SA) inference, which was found to work better than TRWS. For 43 instances the LPQP algorithm obtained better solutions than the previously best known solutions. Figure 4 visualizes some of the instances where the LPQP algorithm leads to a better solution. We observed that the SA solutions seem to hallucinate too much regularity which is not supported by the underlying energy. The scoring of the three algorithms is as follows. LPQP-U: 0.84, SA: 0.74 and TRWS: 0.21. We failed to apply MPLP as the tightening operation did not succeed.

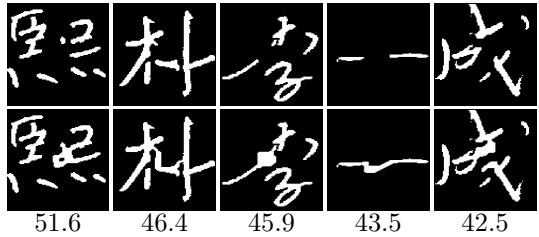

| 51.6 | 46.4 | 45.9 | 43.5 | 42.5 |

*Figure 4.* Results for the Chinese character inpainting dataset. *Top*: results obtained by LPQP-U. *Middle*: solutions from (Nowozin et al., 2011) obtained by simulated annealing. *Bottom*: Energy difference between the simulated annealing solution and the LPQP solution, the larger the value is, the better the LPQP solution is.

## 6. Conclusions

We introduce a novel formulation for MAP inference in graphical models, combining the LP and QP relaxation terms through a KL divergence measure. The resulting problem, albeit being non-convex, gives rise to efficient algorithms built upon known LP solvers.

## Acknowledgments

We would like to thank Joachim Buhmann, Andreas Krause, Cheng Soon Ong and Christian Sigg for insightful discussions. The work was partially supported by the NCCR-MICS, a the Swiss National Science Foundation center, under grant #51NF40-111400.

## References

Beck, A and Teboulle, M. A fast iterative shrinkage-thresholding algorithm for linear inverse problems. *SIAM Journal on Imaging Sciences*, 2(1), 2009.

Bertsekas, D P. *Nonlinear Programming*. Athena Scientific, Belmont, MA, 1999.

Domke, J. Dual decomposition for marginal inference. In *AAAI*, 2011.

Hazan, T and Shashua, A. Norm-product belief propagation: Primal-dual message-passing for approximate inference. *IEEE TIT*, 56(12):6294–6316, 2010.

Jojic, V, Gould, S, and Koller, D. Accelerated dual decomposition for MAP inference. In *ICML*, 2010.

Kappes, J and Schnoerr, C. MAP-Inference for Highly-Connected Graphs with DC-Programming. In *DAGM*, 2008.

Kolmogorov, V. Convergent tree-reweighted message passing for energy minimization. *PAMI*, 2006.

Komodakis, N, Paragios, N, and Tziritas, G. MRF optimization via dual decomposition: Message-passing revisited. In *In ICCV*, 2007.

Kumar, A and Zilberstein, S. Message-Passing Algorithms for Quadratic Programming Formulations of MAP Estimation. In *UAI*, 2011.

Kumar, A, Zilberstein, S, and Toussaint, M. Message-Passing Algorithms for MAP Estimation Using DC Programming. In *AISTATS*, 2012.

Nesterov, Y. A method of solving a convex programming problem with convergence rate $o(1/k^2)$. *Soviet. Math. Dokl.*, 27:372–376, 1983.

Nowozin, S, Rother, C, Bagon, S, Sharp, T, Yao, B, and Kohli, P. Decision tree fields. In *ICCV*, pp. 1668–1675, 2011.

Ravikumar, P and Lafferty, J. Quadratic Programming Relaxations for Metric Labeling and Markov Random Field MAP Estimation. In *ICML*, 2006.

Ravikumar, P, Agarwal, A, and Wainwright, M J. Message-passing for graph-structured linear programs: Proximal methods and rounding schemes. *J. Mach. Learn. Res.*, 11:1043–1080, 2010.

Savchynskyy, B, Kappes, J H, Schmidt, S, and Schnörr, C. A study of Nesterov's scheme for Lagrangian decomposition and MAP labeling. In *CVPR*, pp. 1817–1823, 2011.

Schlesinger, M I. Syntactic analysis of two-dimensional visual signals in noisy conditions. *Kibernetika*, 4: 113–130, 1976.

Sontag, D, Meltzer, T, Globerson, A, Weiss, Y, and Jaakkola, T. Tightening LP relaxations for MAP using message-passing. In *UAI*, pp. 503–510, 2008.

Wainwright, M J and Jordan, M I. *Graphical Models, Exponential Families, and Variational Inference*. 2008.

Yanover, C, Meltzer, T, and Weiss, Y. Linear programming relaxations and belief propagation – an empirical study. *JMLR*, 2006.

Yuille, A L and Rangarajan, A. The concave-convex procedure. *Neural Computation*, 2003.